\documentclass{esannV2}
\usepackage{amsmath}    
\usepackage{amsthm}     
\usepackage{amssymb}    

\newtheorem{theorem}{Theorem}[section] 

\usepackage{subcaption}

\usepackage{graphicx} 
\usepackage{float}
\usepackage{algorithm}
\usepackage{algpseudocode}

\usepackage{physics}

\begin{document}

\title{Fractional Order Distributed Optimization}

\author{Andrei Lixandru$^1$,  Marcel van Gerven$^1$ and S\'ergio Pequito$^2$
%
\thanks{This work was performed during a fellowship supported by the ELLIS Unit Nijmegen.}
%
\vspace{.3cm}\\
%
1- Department of Machine Learning and Neural Computing, \\ Donders Institute, Radboud University, The Netherlands
%
\vspace{.1cm}\\
2- Department of Electrical and Computer Engineering, \\ Instituto Superior Técnico, University of Lisbon, Portugal
}

\maketitle

\begin{abstract}
Distributed optimization is fundamental to modern machine learning applications like federated learning, but existing methods often struggle with ill-conditioned problems and face stability-versus-speed tradeoffs. We introduce fractional order distributed optimization (FrODO); a theoretically-grounded framework that incorporates fractional-order memory terms to enhance convergence properties in challenging optimization landscapes. Our approach achieves provable linear convergence for any strongly connected network. Through empirical validation, our results suggest that FrODO achieves up to 4$\times$ faster convergence versus baselines on ill-conditioned problems and 2-3$\times$ speedup in federated neural network training, while maintaining stability and theoretical guarantees. 
\end{abstract}

\section{Introduction}Distributed optimization (DO) has emerged as a crucial paradigm in machine learning, enabling decentralized computation across networked systems for applications ranging from federated learning to multi-agent coordination \cite{konecny2015federated, lobel2008distributed}. In DO, a network of $N$ agents connected through a directed communication graph $G=(V,E)$ collaboratively solves
\begin{equation}\label{eq:objectiveDO}
x^* = \arg\min\limits_{x \in \mathbb{R}^n} \sum_{i=1}^{N} f_i(x) \,,
\end{equation}
where each agent $i$ maintains a private objective function $f_i(x)$ and iteratively exchanges estimates with its neighbors to converge on a common solution $x^* \in \mathbb{R}^n$. Specifically, agent $i$ communicates with its out-neighbors $\mathcal N_i^+ \subset V$ and updates its estimate based on information from its in-neighbors $\mathcal N_i^- \subset V$, where the existence of paths between all pairs of agents (strong connectivity of $G$) ensures information can flow throughout the network.

A significant challenge in DO arises when objective functions have ill-conditio-ned Hessian matrices, leading to poor convergence rates and instability \cite{yang2019survey}. Recent work in centralized optimization has shown that fractional calculus methods, which incorporate long-term memory, can effectively stabilize optimization trajectories for such challenging objectives \cite{chatterjee2021neo}. Building on this insight, we propose a novel distributed optimization framework that leverages fractional-order memory terms to enhance convergence stability and speed, particularly for problems with adverse curvature properties.

\section{FrODO: Fractional order distributed optimization}

We propose fractional order distributed optimization (FrODO), a novel algorithm that extends traditional proportional-integrative controllers to the distributed setting by incorporating fractional-order memory terms \cite{zamani2007fopid, merrikh2015discrete, yang2019survey}. As outlined in Algorithm 1, FrODO operates in three stages: (1) computation of descent directions using gradient and memory terms 
$
M_i^{(k)} = \sum_{n=1}^{T} \mu(n;\lambda) \cdot g_i^{(k-n)},
$ weighted by $ \mu(n;\lambda) = \flatfrac{\mu_0(n;\lambda)}{\max_n \mu_0(n;\lambda)}$, where $\mu_0(n;\lambda) = \frac{1}{\Gamma(\lambda)} \cdot \frac{1}{n^{1 - \lambda}}$ encapsules a power-law decaying function with respect to $\lambda\in\mathbb{R}^+_0$ -- the fractional-order exponent -- that weights the past gradient terms based on their distance in time from the current iteration; (2) local state updates based on these directions  as $
x_i = x_i - \alpha g_i - \beta M_i,
$
where the gradient term and memory feedback term are parameterized by $\alpha \in \mathbb{R}^n$ and $\beta \in \mathbb{R}^n$; and (3) consensus-driven state alignment across neighboring agents.

\begin{algorithm}[H]
\begin{small}
\caption{FrODO: Fractional order distributed optimization}\label{alg: FrODO}
\begin{algorithmic}\Require 
\State Number of communication rounds $K$; In-neighbors of each agent $\mathcal{N}_i^-$ for $1 \leq i \leq N$; Initial states $\textbf{x} = \{x_1,\ldots,x_N\}$; Private objective functions $\textbf{f}= \{f_1,\ldots,f_N\}: \{\mathbb{R}^n \rightarrow \mathbb{R}\}^N$; Gradient term magnitude$\alpha \in \mathbb{R}^+$; Memory feedback magnitude $\beta \in \mathbb{R}^+$; Memory length of memory feedback $T \in \mathbb{N}^+$; Fractional order exponent $\lambda \in (0,1)$. 
\For{$k$ in $\{1, \ldots, K\}$}
\If{$k>1$}
   \State $g_i\gets \nabla f_i(x_i)$, $i \in \{1, \ldots, N\}$ \Comment{Gradient computation}
   \State $M_i \gets \sum\limits_{n=1}^{T} \mu(n;\lambda) \cdot g_i^{(k-n)}$, $i \in \{1, \ldots, N\}$ \Comment{Memory feedback}
   \State $x_i \gets x_i - \alpha g_i - \beta M_i$, $i \in \{1, \ldots, N\}$ \Comment{Gradient descent with memory}
\EndIf
\State $x_i \gets \frac{1}{|\mathcal{N}_{i}^-|}\sum\limits_{j\in\mathcal{N}_{i}^-}x_j$ , $i \in \{1, \ldots, N\}$ \Comment{Alignment of network agents' states}
\EndFor
\end{algorithmic}
\end{small}
\end{algorithm}

\subsection{Convergence Analysis}

We establish the convergence properties of FrODO through the following theorem:
\begin{theorem}[Convergence of FrODO]\label{theoremConvergence}
Consider a network of \(N\) agents with a strongly connected, directed communication graph \(G\). Let each agent \(i\) have a local objective function \(f_i(x)\) that is \(\mu\)-strongly convex and \(L\)-smooth. For appropriate choices of parameters \(\alpha\), \(\beta\), and \(T\), and fractional order \(\lambda \in (0,1)\), the FrODO algorithm converges to the optimal solution
\[
x^\star = \arg\min_{x \in \mathbb{R}^n} \sum_{i=1}^N f_i(x)
\]
with a linear convergence rate \(O(\rho^k)\), where \(\rho = \max\{|1 - \alpha\mu|, |1 - \alpha L|\} \cdot (1 + \beta C(\lambda)) < 1\), and \(C(\lambda)\) is a \(\lambda\)-dependent constant characterizing the impact of the fractional memory terms.
\end{theorem}
\begin{proof}
We analyze the convergence of the FrODO algorithm by decomposing it into optimization and consensus components, following the approach in \cite{han2019systematic}.
First, consider the local optimization update for agent \(i\): 
$
x_i^{k+1} = x_i^k - \alpha g_i^k - \beta M_i^k
$,
where \(g_i^k = \nabla f_i(x_i^k)\) and \(M_i^k = \sum_{n=1}^T \mu(n;\lambda) g_i^{k-n}\).
Using the \(\mu\)-strong convexity and \(L\)-smoothness of \(f_i\), we apply the Lyapunov function analysis as in \cite{taylor2018lyapunov}. Define the Lyapunov function \(V_i^k = \|x_i^k - x^\star\|^2\) and  we have
$
V_i^{k+1} = \|x_i^k - x^\star - \alpha g_i^k - \beta M_i^k\|^2.
$ 
Expanding and simplifying, we obtain an inequality that relates \(V_i^{k+1}\) to \(V_i^k\) and \(V_i^{k-1}\), incorporating the effects of the memory term \(M_i^k\).
The fractional-order memory term \(M_i^k\) introduces historical gradient information with power-law decay. Its contribution can be bounded using properties of the memory weights \(\mu(n;\lambda)\), ensuring that the optimization error decreases at a linear rate. Appropriate choices of \(\alpha\) and \(\beta\) ensure that the contraction factor \(\rho < 1\), following the methodology in \cite{taylor2018lyapunov}.
For the consensus component, agents communicate over a strongly connected graph \(G\). The consensus error decreases geometrically due to the properties of the communication protocol, as established in \cite{olfati2004consensus}. Specifically, the disagreement among agents shrinks at a rate determined by the second-largest eigenvalue of the communication matrix, resulting in a contraction factor \(\sigma < 1\).
By combining the optimization and consensus analyses through the decomposition approach in \cite{han2019systematic}, we conclude that the overall error satisfies:
\[
\|x_i^k - x^\star\| \leq C \cdot \max\{\rho, \sigma\}^k \,,
\]
where \(C\) depends on initial conditions. This shows that the FrODO algorithm converges linearly to \(x^\star\), with the convergence rate determined by the worse of the two contraction factors \(\rho\) and \(\sigma\).
\end{proof}

Notice that for any \(\lambda \in (0,1)\), we can choose \(\alpha\), \(\beta\) to ensure \(\rho < 1\), yielding linear convergence. The optimal \(\lambda\) balances short-term adaptivity (\(\lambda \to 0\)), long-term stability (\(\lambda \to 1\)), and memory efficiency (affecting \(C(\lambda)\)), which explains why moderate \(\lambda\) values (0.1-0.2) often perform well empirically.  This improved theoretical understanding helps explain the empirical advantages observed in the ill-conditioned test cases and provides guidance for practical implementation.

\subsection{Complexity Analysis}

The computational complexity of FrODO can be analyzed per iteration as follows:
\begin{theorem}[Computational Complexity]
For a network of \(N\) agents optimizing \(n\)-dimensional variables with memory length \(T\), each iteration of the FrODO algorithm requires per agent \(O(n)\) computation for gradient evaluation, \(O(Tn)\) computation for memory term calculation, \(O(d_i n)\) communication with neighbors (where \(d_i\) is the degree of agent \(i\)), and \(O(Tn)\) space to store historical gradients. The total communication complexity per iteration is \(O(|E| n)\), where \(|E|\) is the number of edges in the communication graph.
\end{theorem}

\begin{proof}
At each iteration, each agent \(i\) evaluates its gradient \(\nabla f_i(x_i)\), requiring \(O(n)\) computations. It then calculates the memory term using \(T\) previous gradients, which takes \(O(Tn)\) computations. The local estimate update is an \(O(n)\) operation. Communication with neighbors involves exchanging information, which has a complexity of \(O(d_i n)\) for agent \(i\).
Summing over all agents, the total communication complexity per iteration is \(O(|E| n)\), since the sum of all agents' degrees \(\sum_{i=1}^N d_i = |E|\) in directed graphs (considering out-degrees). The space complexity per agent is \(O(Tn)\) due to storing \(T\) historical gradients of dimension \(n\).
\end{proof}

The additional memory overhead of \(O(Tn)\) per agent is justified by the improved convergence properties. In problems with ill-conditioned Hessians, traditional methods may require significantly more iterations to achieve similar accuracy, making the FrODO algorithm more efficient overall.
In the next section, we illustrate its performance in a variety of settings and sensitivity to different parameter values and initial conditions.

\section{Experiments}
In the following subsections, we examine two experiments: a pedagogical example using a quadratic function with an ill-conditioned Hessian, and an application-driven experiment in federated learning, where we train an artificial neural network across multiple agents. Although Theorem~\ref{theoremConvergence} establishes convergence for any strongly connected directed graph, our experiments focus on fully connected networks with optimal communication weights as defined in \cite{optimized_communication_weights} to provide a clear baseline for performance comparison.

\subsection{Experiment 1: Objective with an ill-defined Hessian}

To demonstrate our algorithm's effectiveness, we consider a DO setting with four agents optimizing objective functions that yield an ill-conditioned global Hessian. The objective functions are given by 
$f_1(x_1, x_2) = 0.5(2-x_1)^2 + 0.005x_2^2$, $f_2(x_1, x_2) = 0.5(2+x_1)^2 + 0.005x_2^2$, $f_3(x_1, x_2) = 0.5x_1^2 + 0.005(2-x_2^2)$ and $f_4(x_1, x_2) = 0.5x_1^2 + 0.005(2+x_2^2)$.
The resulting global objective $f_{global}(x_1, x_2) = x_1^2 + 0.02x_2^2 + 4.04$ has its minimum at $(0,0)$.

\begin{figure}[H]
    \centering
    \begin{subfigure}{.5\textwidth}
        \centering
        \includegraphics[height=0.6\textwidth]{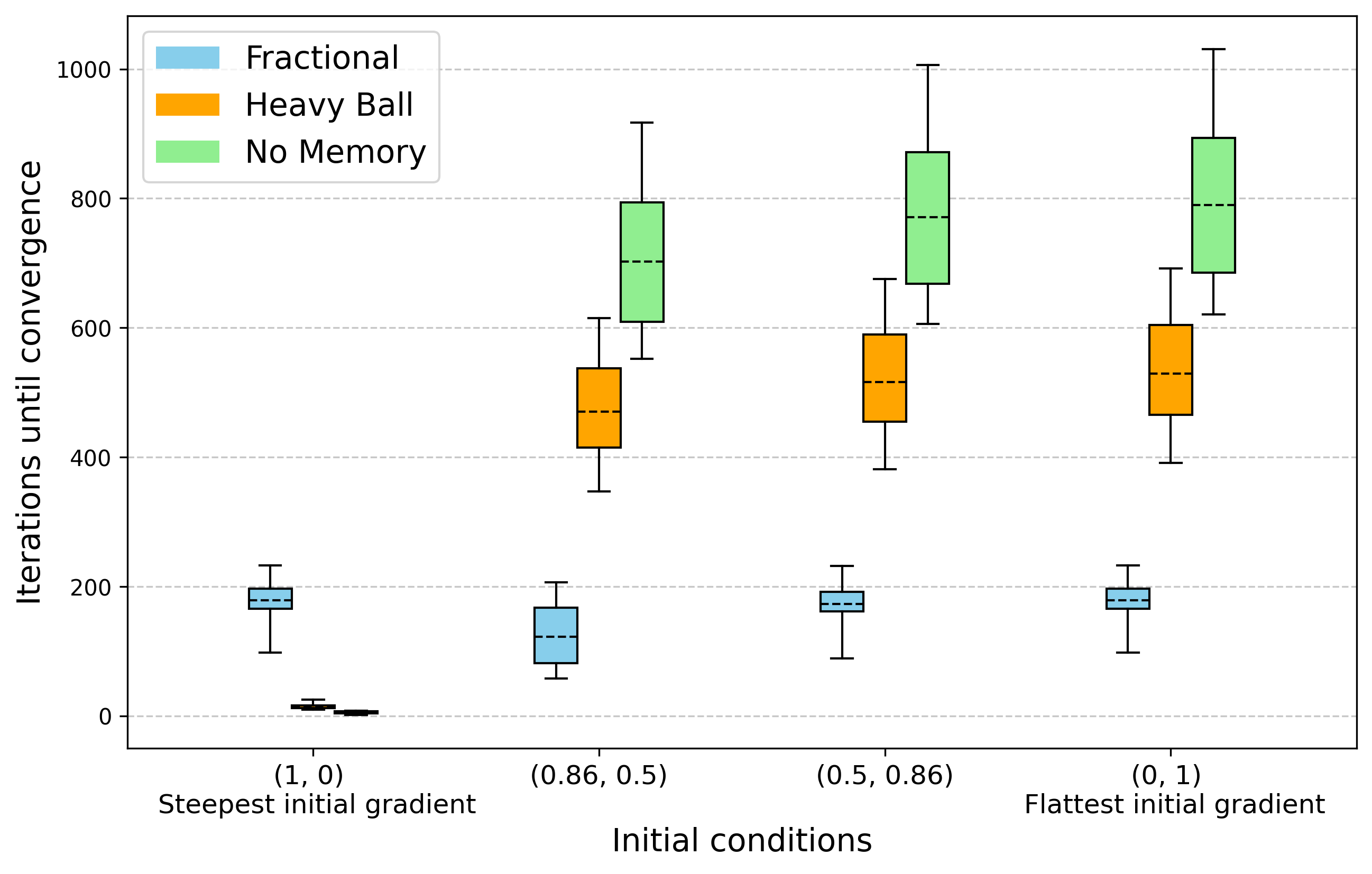}        
    \end{subfigure}%
     \begin{subfigure}{.5\textwidth}
        \centering
        \includegraphics[height=0.6\textwidth]{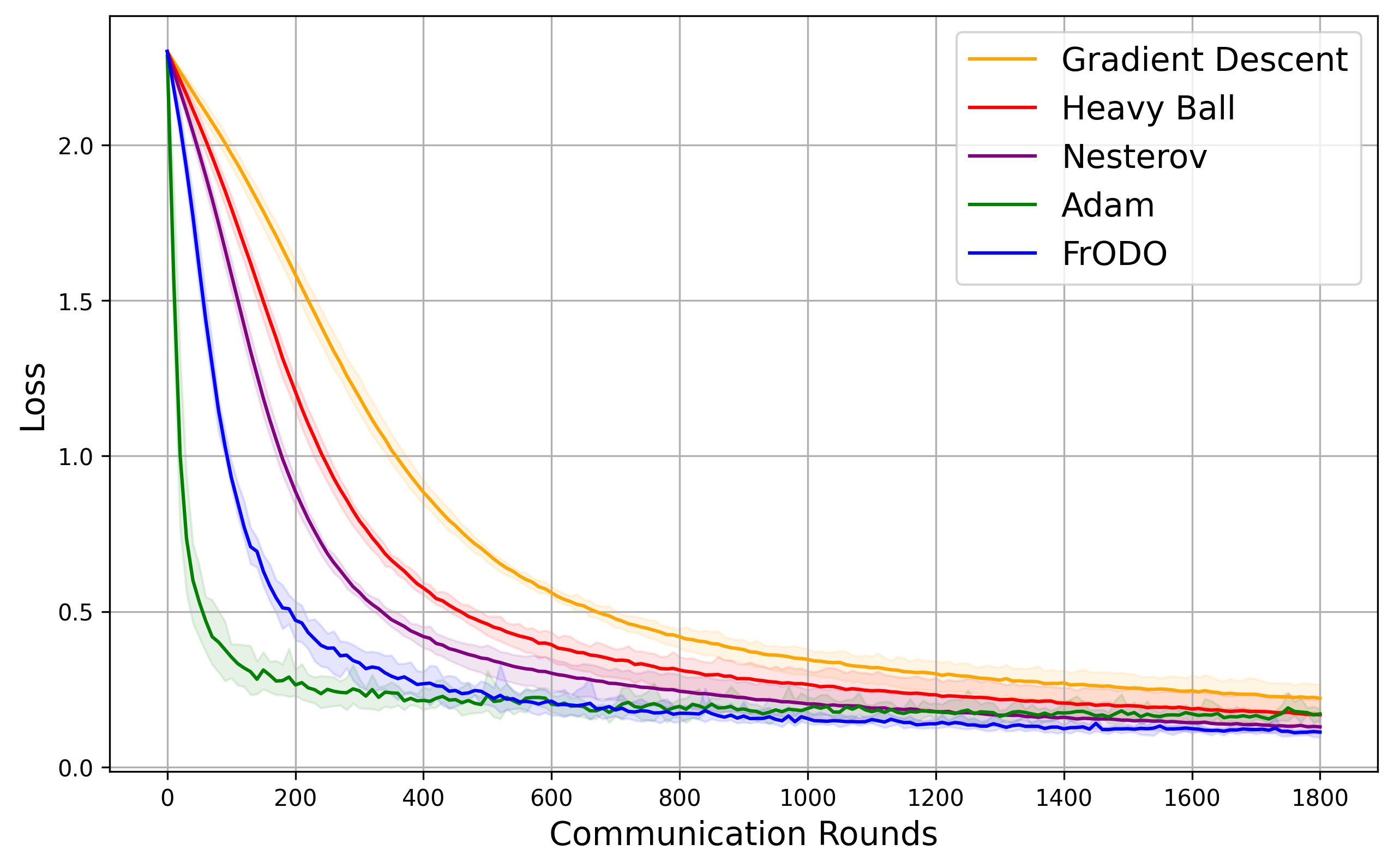}        
    \end{subfigure}%

    \caption{Results for Experiment 1 (left) and Experiment 2 (right).}
    \label{fig:experimentalSetups}
\end{figure}

We evaluate convergence from four unit circle starting points: $(1,0)$, $(0.86,0.5)$, $(0.5,0.86)$, and $(0,1)$, where $(1,0)$ and $(0,1)$ represent the steepest and flattest initial gradients respectively. We compare three variants: Fractional (our proposed long-term memory approach), Heavy Ball ($T=1$), and No Memory ($\beta=0$). Experiments use 100 hyperparameter sets with $\alpha \in [0.6, 1]$, $\beta \in [\alpha/1.5, \alpha/2.5]$, $\lambda \in [0.1, 0.2]$, and $T \in [80, 100]$.

Figure~\ref{fig:experimentalSetups} (left) shows that our approach outperforms the other approaches. Two-sided Kolmogorov-Smirnov tests reveal that our fractional approach maintains consistent performance between steepest and flattest gradients ($p=1$), while other methods show significant performance differences ($p<0.00001$). In additional tests with uniformly sampled initial states on the unit circle, our method demonstrates superior convergence (427$\pm$145 iterations) compared to Heavy Ball (1,538$\pm$400) and No Memory (1,864$\pm$312), with one-sided KS tests confirming significance ($p<0.00001$).





\subsection{Experiment 2: Artificial neural networks}
We also evaluate FrODO in a federated learning setting using two ANNs (918,192 parameters each) for MNIST \cite{lecun1998mnist} image classification. Each agent receives a distinct balanced dataset and optimizes network weights using mini-batch loss (batch size 64). We compare against standard baselines: gradient descent, Nesterov momentum, heavy ball ($T=1$), and Adam, implemented as variations of Algorithm~\ref{alg: FrODO} by modifying the stage 2 descent terms.


Figure~\ref{fig:experimentalSetups} (right) shows that, across five runs with randomized initializations and data partitions, FrODO demonstrates faster convergence than most baselines while maintaining comparable final performance to Adam.

\section{Discussion}
Our results indicate that FrODO effectively addresses fundamental challenges in distributed optimization, particularly for ill-conditioned problems. The power-law memory decay provides theoretical advantages through balanced historical and current gradient information, though memory requirements of $O(Tn)$ warrant consideration for high-dimensional problems. Our empirical findings suggest setting $\lambda \in [0.1, 0.2]$ with $T \geq 80$ for most applications, while larger $\lambda$ values benefit more ill-conditioned problems. 

Future research directions include adaptive parameter tuning, non-convex theoretical analysis, memory-efficient implementations, and extensions to time-varying networks. While our current results focus on distributed optimization, the principles of fractional-order memory could impact broader areas including multi-agent reinforcement learning and distributed control systems, providing a foundation for addressing emerging challenges in modern machine learning systems.

\begin{footnotesize}
\bibliographystyle{unsrt}
\bibliography{references_final}
\end{footnotesize}

\end{document}